%% file: acl_latex.tex
\documentclass[11pt]{article}

\usepackage[final]{acl}

\usepackage{times}
\usepackage{latexsym}

\usepackage[T1]{fontenc}

\usepackage[utf8]{inputenc}

\usepackage{microtype}

\usepackage{inconsolata}

\usepackage{graphicx}
\usepackage{booktabs}   
\usepackage{multirow}   
\usepackage{amsmath}
\usepackage{amsfonts}
\usepackage{amssymb} 
\usepackage{makecell}   
\usepackage[table]{xcolor}

\usepackage{tabularx}   
\usepackage{caption}    
\usepackage{longtable}      
\usepackage{lipsum}                       
\usepackage{enumitem}                     
\usepackage{minted}
\usepackage[most]{tcolorbox}
\usepackage[utf8]{inputenc}               

\definecolor{PosGreen}{RGB}{0, 150, 0} 
\definecolor{NegBlue}{RGB}{0, 0, 255}  

\definecolor{tablehead}{gray}{0.9}
\definecolor{darkgreen}{rgb}{0.0, 0.5, 0.0}
\definecolor{darkred}{rgb}{0.6, 0.0, 0.0}

\newcommand\footnoteONLYtext[1]{
    \let \mybackup \thefootnote
    \let \thefootnote \relax
    \footnotetext{#1}
    \let \thefootnote \mybackup
    \let \mybackup \imareallyundefinedcommand
}
\title{Learning to Evolve: A Self-Improving Framework for Multi-Agent Systems via Textual Parameter Graph Optimization}

\author{
\textbf{Shan He}$^{*}$ \quad
\textbf{Runze Wang}$^{*}$ \quad
\textbf{Zhuoyun Du} \quad
\textbf{Huiyu Bai} \\
\textbf{Zouying Cao} \quad
\textbf{Yu Cheng} \quad
\textbf{Bo Zheng}$^{\dagger}$ \\
[3pt]
Future Living Lab of Alibaba \\
[3pt]
\texttt{\{shanhe.hs, yunze.wrz\}@alibaba-inc.com}
}

\begin{document}
\maketitle
\footnoteONLYtext{$^*$Equal contribution.}
\footnoteONLYtext{$^\dagger$Corresponding author.}
\begin{abstract}
Designing and Optimizing multi-agent systems (MAS) is a complex, labor-intensive process of "Agent Engineering." Existing automatic optimization methods, primarily focused on \textbf{flat prompt tuning}, lack the structural awareness to debug the intricate web of interactions in MAS. More critically, these optimizers are \textbf{static}; they do not learn from experience to improve their own optimization strategies. To address these gaps, we introduce \textbf{Textual Parameter Graph Optimization (TPGO)}, a framework that enables a multi-agent system to \textbf{learn to evolve}. TPGO first models the MAS as a \textbf{Textual Parameter Graph (TPG)}, where agents, tools, and workflows are modular, optimizable nodes. To guide evolution, we derive \textbf{"textual gradients"}, structured natural language feedback from execution traces, to pinpoint failures and suggest granular modifications. The core of our framework is \textbf{Group Relative Agent Optimization (GRAO)}, a novel \textbf{meta-learning strategy} that learns from historical optimization experiences. By analyzing past successes and failures, GRAO becomes progressively better at proposing effective updates, allowing the system to \textbf{learn how to optimize itself}. Extensive experiments on complex benchmarks like GAIA and MCP-Universe show that TPGO significantly enhances the performance of state-of-the-art agent frameworks, achieving higher success rates through automated, \textbf{self-improving optimization}.
\end{abstract}

\section{Introduction}

\begin{figure}
    \setlength{\abovecaptionskip}{5 pt} 
    \setlength{\belowcaptionskip}{-15pt} 
    \centering
    \includegraphics[width=1.0\linewidth]{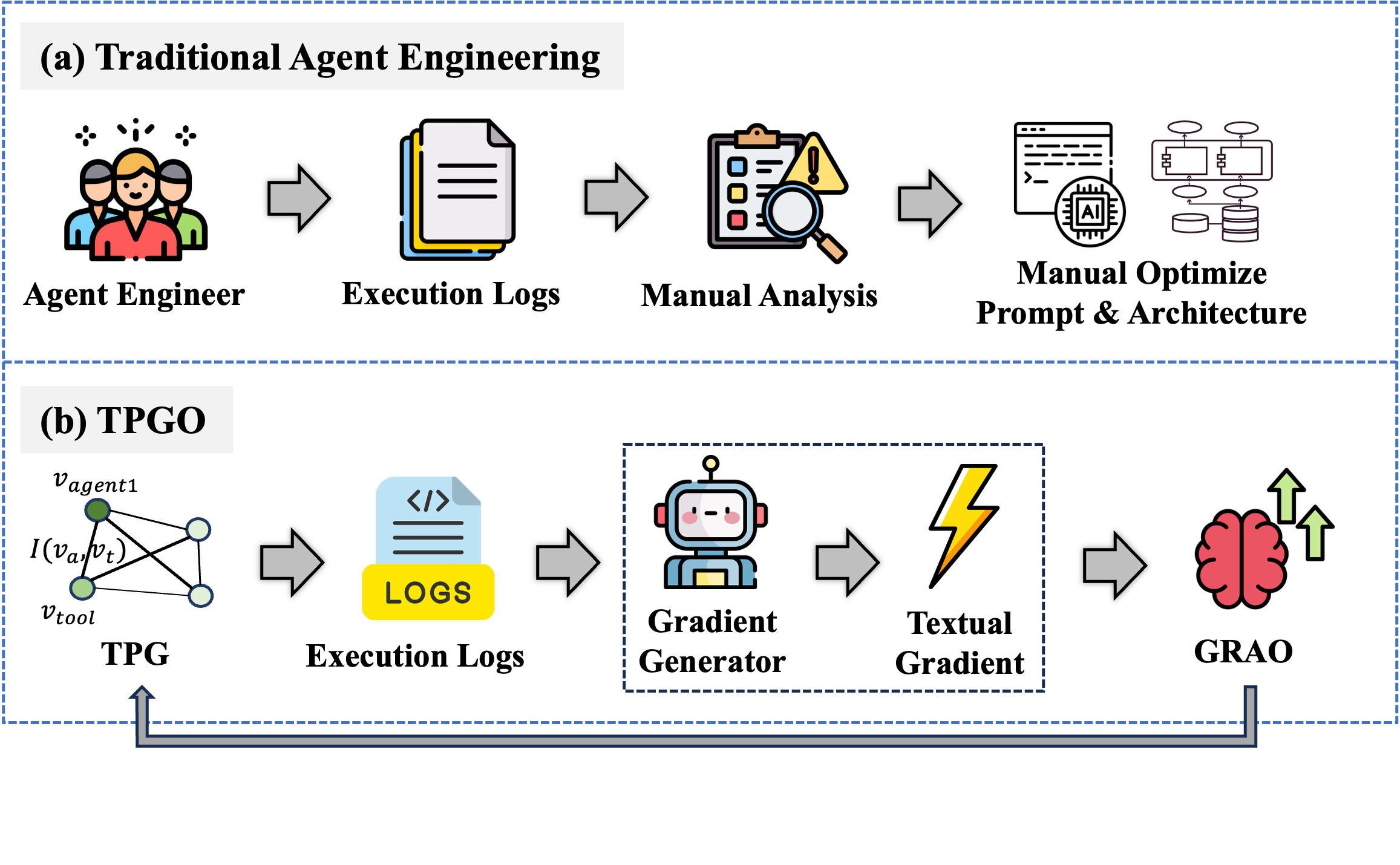}
    \caption{The pipelines of Traditional Agent Engineering and the proposed TPGO in multi-agent system optimization.}
    \label{fig:intro}
\end{figure}

The rapid advancement of Large Language Models (LLMs) has catalyzed a paradigm shift in artificial intelligence, moving from passive question-answering interfaces to autonomous agents capable of tool use and decision-making. This evolution has recently culminated in the development of \textbf{Multi-Agent Systems (MAS)}, where diverse agents collaborate to solve complex, multi-step problems ranging from software engineering to open-ended reasoning \cite{DBLP:conf/uist/ParkOCMLB23, DBLP:conf/ijcai/GuoCWCPCW024, hong2023metagpt}. By decomposing intricate tasks into specialized sub-routines, MAS have demonstrated capabilities that significantly surpass solitary LLMs.

However, the efficacy of these systems is critically dependent on the precise configuration of their textual components. An "Agent Engineer" must meticulously craft not only the system prompts for individual agents but also the descriptions of tools, the protocols for inter-agent communication, and the overarching workflow \cite{talebirad2023multi, luo2025large, tran2025multi}. This process, often termed \textit{Agent Engineering}, represents a high-dimensional and unstructured optimization challenge. As illustrated in Figure \ref{fig:intro}(a), the opaque and non-deterministic nature of MAS interactions makes manual tuning a labor-intensive trial-and-error process that is difficult to scale and rarely converges to an optimal state. This establishes the \textbf{automatic optimization of MAS} as a crucial, yet unsolved, frontier. 

In real-world applications, this optimization challenge manifests in two primary forms. On the one hand, \textbf{exploratory optimization} addresses settings where the system must improve without a "golden" answer, relying on self-correction from execution feedback. On the other hand, \textbf{imitative optimization} seeks to align the system's behavior with desired outcomes by learning from expert demonstrations or ideal solutions.

To alleviate this manual burden, the first wave of research has explored Automatic Prompt Optimization (APO) \cite{wang2023promptagent, yuksekgonul2024textgrad}. While promising, these methods fall short in addressing the unique complexities of MAS in two fundamental ways. First, they primarily focus on "flat" prompt optimization and lack the \textbf{structural awareness} required to navigate the intricate web of interactions within a multi-agent architecture. A system failure often stems not from a single flawed prompt, but from a subtle bug in a tool's definition, a logical gap in the workflow, or a misaligned communication protocol, components that current APO methods are ill-equipped to identify and correct \cite{fang2025comprehensive}. Second, and more fundamentally, existing optimizers are \textbf{static}; they do not learn from the optimization process itself. They execute a search or apply a gradient-like update for a given problem, but they lack the mechanism to internalize past failures and successes to become better optimizers over time. This exposes a critical pain point: for MAS to achieve true autonomy, \textbf{the optimization system itself must learn to evolve}. This forms the core motivation for our work: to create an optimizer that not only refines the agent system but also refines its own optimization strategy through experience.

To bridge these gaps, we propose \textbf{Textual Parameter Graph Optimization (TPGO)}, a novel framework that re-conceptualizes MAS optimization as a \textbf{graph evolution problem}, as depicted in Figure \ref{fig:intro}(b). To address the structural challenge, we first formalize the system's configuration as a \textbf{Textual Parameter Graph (TPG)}. In this representation, agents, tools, and logical units are modular nodes, and their interactions are directed edges. This structured view allows us to move beyond flat text editing to perform targeted, structural modifications on the system's architecture.

Crucially, to enable the optimizer to evolve, we introduce \textbf{Group Relative Agent Optimization (GRAO)}, a \textbf{meta-learning mechanism} that acts as the "brain" of the optimization process. By clustering historical error patterns and retrieving successful past optimization strategies, GRAO learns to generate more effective and targeted update proposals over time. It uses \textbf{Textual Gradients}, structured natural language feedback from execution traces, as its input signal, allowing it to reason about and correct semantic errors within the graph. In essence, GRAO empowers our framework to \textbf{"learn how to optimize."}

Our contributions can be summarized as follows:
\begin{itemize}
\item We propose \textbf{TPGO}, the first framework to treat multi-agent system optimization as a \textbf{graph evolution problem}, enabling both structural and semantic refinement of complex agent architectures.
\item We introduce the \textbf{Textual Parameter Graph (TPG)}, a structured representation that disentangles monolithic textual configurations into modular, individually optimizable semantic units.
\item We design \textbf{Group Relative Agent Optimization (GRAO)}, a novel \textbf{self-evolving meta-optimization strategy} that leverages historical experience to continuously improve the quality of its optimization proposals.
\item Extensive experiments demonstrate that our framework significantly enhances the performance and efficiency of state-of-the-art (SOTA) agent systems on complex benchmarks, achieving higher success rates through \textbf{automated, evolutionary optimization}.
\end{itemize}

\section{Related Work}

\subsection{Language Agent Systems}
While single-agent systems (SAS) such as the ReAct framework \cite{DBLP:conf/iclr/YaoZYDSN023} offer simplicity and efficiency, their capabilities are inherently limited. To overcome these limitations, LLM-based multi-agent systems (MAS) have emerged as a powerful paradigm for complex problem-solving. The prevailing design methodology involves decomposing a task into sub-tasks and assigning them to specialized agents that emulate collaborative human teams \citep{DBLP:conf/uist/ParkOCMLB23, DBLP:conf/ijcai/GuoCWCPCW024}. This approach, however, introduces a vast design space with numerous critical choices that shape collaboration and behavior. For instance, at the system level, designers must define the collaboration structure, choosing from predefined workflows like hierarchical organizations \citep[e.g., MaCTG;][]{zhao2024mactg}, iterative refinement, or multi-agent debate. At the agent level, their behavior is further guided by manually crafted personas \citep{lin2025creativity}, which is crucial for achieving sophisticated outcomes beyond single-agent performance. Consequently, the overall system performance is highly sensitive to this vast array of design choices, including the composition of the agent team, the collaboration mechanism, the communication protocol, and the specific configuration of each agent's persona and proactivity \citep{DBLP:conf/acl/ZhangX0LHD24,lin2025creativity}.

\subsection{Automatic Prompt Optimization} 
Recent efforts in automatic prompt optimization have pursued three primary directions. Search-based methods, such as PromptAgent \cite{wang2023promptagent}, leverage MCTS for strategic exploration. Evolutionary algorithms like EvoPrompt \cite{tong2025evoprompt} iteratively evolve prompts to select for high-performing candidates. In contrast, gradient-based approaches such as TextGrad \cite{yuksekgonul2024textgrad} frame prompts as differentiable parameters, optimizing them through backpropagation.

A shared limitation of these methods is their focus on textual content over system architecture. They optimize individual prompts in isolation but are ill-equipped to resolve failures originating from the complex interplay of agents, tools, and logic in a Multi-Agent System. Lacking the ability to modify the system's interaction structure, current APO techniques leave core architectural problems unsolved. Our work directly confronts this challenge by introducing a structurally-aware optimization framework.

\section{Task Definition}
\label{sec:task_definition}

The central problem this paper addresses is the automated optimization of Multi-Agent Systems (MAS). The performance of an MAS is dictated by its configuration, a complex collection of textual components that includes agent prompts, tool descriptions, and interaction protocols. The manual process of tuning these components, known as Agent Engineering, is a laborious and unsystematic endeavor. The high dimensionality and non-deterministic nature of LLM interactions make this optimization landscape exceedingly difficult to navigate, with no guarantee of convergence to an optimal state.

\subsection{Problem Formulation}
We formalize a multi-agent system $\mathcal{A}$ as being parameterized by a collection of configurable natural language elements, which we term \textbf{textual parameters} $\Theta$. These parameters encompass all the system's textual components, such as system prompts for each agent $\{P_i\}$, descriptions of available tools $\{D_j\}$, and the rules governing inter-agent communication.

Given a downstream task distribution $\mathcal{T}$, represented by a dataset of input-output pairs $(Q, A) = \{(q_k, a_k)\}_{k=1}^N$, the system $\mathcal{A}(\Theta)$ takes an input query $q_k$ and produces a final output $\hat{a}_k$. This execution generates a corresponding trajectory $\tau_k$, which is a detailed record of the agents' intermediate reasoning steps, tool calls, and communications.

The objective is to discover the optimal set of textual parameters $\Theta^*$ that maximizes the system's performance as measured by a reward metric $\mathcal{R}$ (e.g., success rate). This optimization problem can be expressed as:
\begin{equation}
\label{eq:main_opt}
\Theta^* = \arg\max_{\Theta \in \mathcal{S}} \mathbb{E}_{(q,a) \sim \mathcal{T}}[\mathcal{R}(\mathcal{A}(\Theta, q), a)]
\end{equation}
where $\mathcal{S}$ represents the vast, discrete, and unstructured space of all possible textual configurations. The intractability of navigating $\mathcal{S}$ is the primary obstacle, as conventional optimization methods and exhaustive search are infeasible. Our work aims to automate this discovery process, creating a system that can methodically and efficiently explore this space to find high-performing configurations. The proposed framework, Textual Parameter Graph Optimization (TPGO), is designed to impose structure on this problem and enable a learning-based search for $\Theta^*$, as detailed in the following section.

\section{Methodology}
\label{sec:methodology}

\begin{figure*}
    \setlength{\abovecaptionskip}{2pt} 
    \setlength{\belowcaptionskip}{-12pt} 
    \centering
    \includegraphics[width=1\linewidth]{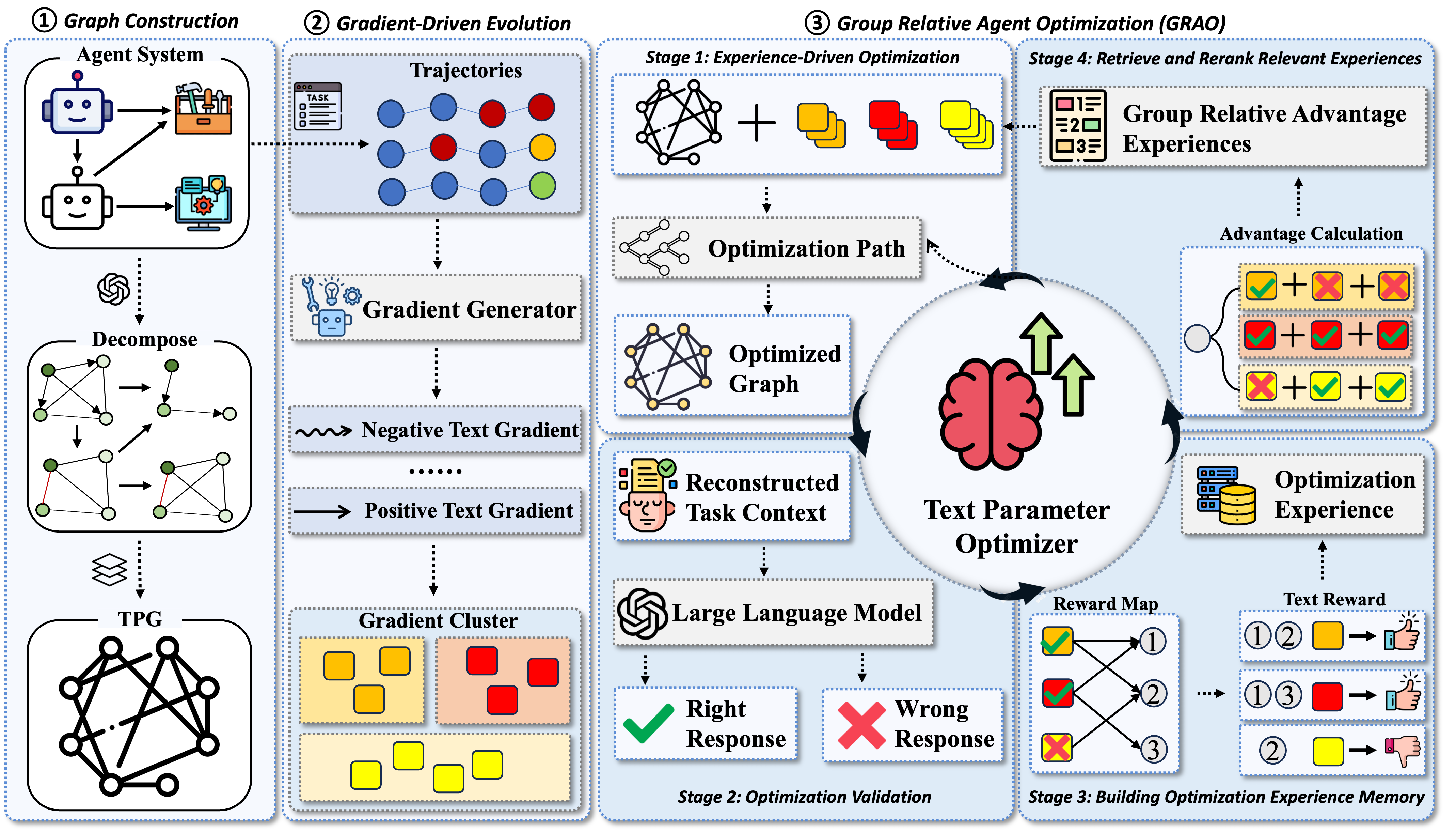}
    \caption{\textbf{Overview of the Textual Parameter Graph Optimization (TPGO) framework.} The framework operates in a closed-loop cycle: (1) \textit{Graph Construction}, where unstructured agent prompts are disentangled into a structured \textbf{Textual Parameter Graph (TPG)}; (2) \textit{Gradient-Driven Evolution}, which generates semantic \textbf{Textual Gradients} from execution trajectories to diagnose faults; and (3) \textit{Group Relative Agent Optimization (GRAO)}, a meta-learning module that leverages an optimization experience memory to guide the \texttt{Optimizer LLM} in generating effective graph updates ($\Delta\mathcal{G}$).}
    \label{fig:main_architecture}
\end{figure*}

\subsection{Overview}
To enable autonomous self-evolution in complex multi-agent systems, our framework, \textbf{Textual Parameter Graph Optimization (TPGO)}, reframes the entire system as a structured, optimizable object. We posit that the essence of any LLM-based agent, including its role, logic, and capabilities, is encapsulated within its textual configuration. TPGO operates in a closed-loop cycle designed to systematically refine this configuration, as illustrated in Figure \ref{fig:main_architecture}. The cycle consists of three core phases: (1) \textbf{Graph Construction}, where opaque prompts are decomposed into a transparent and modular Textual Parameter Graph (TPG); (2) \textbf{Gradient-Driven Evolution}, where semantic feedback signals called "textual gradients" are derived from execution traces to guide graph updates; and (3) \textbf{Group Relative Agent Optimization (GRAO)}, a meta-optimization layer that enables the optimizer itself to learn and improve from experience. This holistic design ensures that both the agent system and the optimization process evolve in tandem.

\subsection{Textual Parameter Graph}
\label{subsec:graph_construction}
Standard multi-agent systems often rely on monolithic, unstructured textual prompts to define agent behaviors, an approach that is brittle, difficult to debug, and opaque to systematic optimization. To overcome these limitations, we introduce a novel representation for the system's textual configuration $\Theta$, which we formalize as a directed \textbf{Textual Parameter Graph} (TPG), denoted by $\mathcal{G} = (\mathcal{V}, \mathcal{E})$. This graph structure disassembles the system's logic into modular, inspectable, and individually optimizable components.

\paragraph{Node Representation ($\mathcal{V}$).}
Each node $v_i \in \mathcal{V}$ represents a discrete \textit{semantic unit} of the system. We generate these nodes by hierarchically decomposing the initial system prompts using a dedicated Parser LLM. This process supports a nested graph structure, where high-level agents can form a main graph, while their specific components are modeled as interconnected subgraphs. The set of nodes encompasses three primary types:
\begin{itemize}
    \item \textbf{Role Nodes:} Define an agent's core persona, high-level objectives, and strategic directives.
    \item \textbf{Logic Nodes:} Encapsulate specific reasoning protocols, operational constraints, or chained thought processes.
    \item \textbf{Tool Nodes:} Contain functional descriptions, API specifications, and usage examples for external tools.
\end{itemize}
The complete textual configuration $\Theta$ is constituted by the aggregation of content from all nodes, such that $\Theta = \bigcup_{v_i \in \mathcal{V}} \text{Content}(v_i)$.

\paragraph{Edge Representation and System Dynamics ($\mathcal{E}$).}
The directed edges $\mathcal{E}$ model the dependencies and logical flow between semantic units. An edge $e_{ij}$ from node $v_i$ to $v_j$ signifies that the content of $v_i$ informs, constrains, or provides context for $v_j$. These edges are crucial for defining the system's dynamic behavior by representing relationships such as intra-agent flow connecting a Role Node to a Logic Node, tool integration linking a Logic Node to a Tool Node, and inter-agent communication established by connecting nodes across different agent subgraphs.

\paragraph{Graph-based Optimization.}
This graph-based representation transforms the challenge of prompt engineering into a structured graph optimization problem. It enables fine-grained manipulation of the agent system's behavior through two primary mechanisms. First, through \textbf{node content refinement}, the text within a specific node can be modified for targeted updates, analogous to parameter tuning in traditional models. For example, editing a Tool Node's content can improve an agent's ability to use it correctly. Second, via \textbf{structural modification}, the graph's topology can be altered to dynamically change system capabilities. Pruning an edge to a Tool Node can revoke an agent's access to that tool, while adding new nodes and edges can introduce new skills or agents into the system. This structured approach facilitates not only manual debugging but also automated system evolution.

\subsection{Gradient-Driven Evolution}
\label{subsec:gradient_evolution}
Given the system's representation as a Textual Parameter Graph (TPG) $\mathcal{G}$, our objective is to iteratively refine $\mathcal{G}$ to maximize task performance. Since traditional gradient-based optimization is not applicable to discrete textual structures, we introduce a novel paradigm centered on \textbf{Textual Gradients}. These are structured natural language critiques and suggestions that guide the evolution of the graph through a three-stage process of generation, aggregation, and application.

\paragraph{Step 1: Trajectory Diagnosis and Gradient Generation.}
For a given batch of tasks, the MAS is executed to collect a set of trajectories $\mathcal{T} = \{\tau_1, \dots, \tau_N\}$. Each trajectory $\tau_k$ is a detailed log of the system's reasoning, tool use, and communication. A diagnostic model then analyzes these trajectories, comparing them against ground-truth solutions or ideal behaviors to identify discrepancies. From this analysis, the model generates a textual gradient $\nabla_{\text{text}}$ for each informative trajectory, which provides a rich, interpretable signal for improvement: $\nabla_{\text{text}} = \text{Generate}(\tau_k) \rightarrow \{ \delta^+, \delta^- \}$. This gradient is composed of two parts:
\begin{itemize}
    \item \textbf{Positive Gradient ($\delta^+$):} Extracted from successful trajectories, $\delta^+$ distills high-quality reasoning patterns or effective tool-use strategies. It serves as a template for reinforcing correct behaviors (e.g., "The agent correctly broke down the problem into sub-steps A, B, and C.").
    \item \textbf{Negative Gradient ($\delta^-$):} Generated from failed trajectories, $\delta^-$ pinpoints specific errors like hallucinations or tool misuse and suggests a corrective action (e.g., "The agent hallucinated a parameter for the $search$ tool; it should have used $list\_parameters$ first.").
\end{itemize}

\paragraph{Step 2: Gradient Aggregation via Clustering.}
Applying updates from every individual negative gradient $\delta^-$ can be inefficient and lead to noisy, conflicting modifications. To address this, we aggregate gradients to identify systemic error patterns. We employ a semantic clustering approach on the set of all generated negative gradients $\{\delta^-_k\}$. By embedding the textual descriptions of errors, we group them into clusters, where each cluster represents a recurring type of failure, such as "misunderstanding a specific tool's constraints" or "inefficient inter-agent communication." This step allows the system to target underlying root causes rather than fixing isolated faults.

\paragraph{Step 3: Generating and Applying Optimization Proposals.}
For each error cluster, an \textbf{Optimizer LLM} generates a concrete \textbf{Optimization Proposal}, $\Delta\mathcal{G}$. This proposal is a machine-readable plan to modify the TPG, taking the representative error description from the cluster and the current graph $\mathcal{G}$ as input. It outputs a set of specific graph edit operations, such as:
\begin{itemize}
    \item \texttt{REWRITE\_NODE(v, new\_content)}: Modifies the text of a node.
    \item \texttt{PRUNE\_EDGE(u, v)}: Removes a connection to disable a faulty pathway.
    \item \texttt{ADD\_NODE(v\_new, content)}: Introduces a new skill or constraint.
    \item \texttt{ADD\_EDGE(u, v)}: Establishes a new dependency or logical flow.
\end{itemize}
Finally, the proposal is applied to the graph to produce the evolved version $\mathcal{G}' \leftarrow \mathcal{G} \oplus \Delta\mathcal{G}$, where $\oplus$ denotes the application of the modifications. This iterative cycle drives the evolution of the agent system.

\subsection{Self-Evolving Optimization via GRAO}
\label{subsec:grao}
The gradient-driven evolution process relies on an \texttt{Optimizer LLM} to generate optimization proposals ($\Delta\mathcal{G}$). However, a static optimizer may repeatedly suggest ineffective modifications. We introduce \textbf{Group Relative Agent Optimization (GRAO)}, a meta-optimization framework designed to make the optimization process itself self-improving. The core principle of GRAO is to create a feedback loop where the system learns from the efficacy of its past optimizations to generate better proposals in the future. This is achieved through a cycle of validation, memorization, and adaptive in-context learning.

\paragraph{Optimization Validation and Effectiveness Scoring.}
After an optimization proposal $\Delta\mathcal{G}$ is applied to create an evolved graph $\mathcal{G}'$, its impact is empirically validated. To avoid the high cost of a full re-evaluation, we perform targeted validation: the new agent configuration $\mathcal{A}(\mathcal{G}')$ is run specifically on the subset of tasks that contributed to the error cluster being addressed. By comparing the new outcomes to the previous failures, we compute an \textbf{effectiveness score}, $E(\Delta\mathcal{G}) \in [0, 1]$, which quantifies the proposal's success rate on the problematic task subset.

\paragraph{Building an Optimization Experience Memory.}
Each optimization attempt is recorded as a structured entry in a persistent \textbf{Optimization Experience Memory}. This memory serves as a knowledge base of what has been tried and how well it worked. Each entry is a tuple containing three key components:
\begin{itemize}
    \item \textbf{Problem Context ($C$):} The semantic description of the error cluster that prompted the optimization.
    \item \textbf{Solution Proposal ($\Delta\mathcal{G}$):} The specific set of graph edit operations that were applied.
    \item \textbf{Outcome ($E$):} The resulting effectiveness score from the validation step.
\end{itemize}
Over time, this memory accumulates a rich dataset of problem-solution-outcome triplets, documenting the system's learning journey.

\paragraph{Adaptive Proposal Generation via In-Context Learning.}
The Optimization Experience Memory enables the \texttt{Optimizer LLM} to evolve from a naive reasoner into an expert guided by experience. When a new error cluster with description $C_{\text{new}}$ is identified, the GRAO mechanism initiates a retrieval-augmented generation process. First, it retrieves historical problem contexts $\{C_k\}$ from memory that are semantically similar to the current problem $C_{\text{new}}$, forming a "group" of related issues. Second, these retrieved experiences are ranked by their effectiveness scores $\{E_k\}$, a "group relative" ranking that identifies which past optimization strategies have been most successful for this class of error. Finally, the highest-ranked exemplars are formatted as few-shot examples and prepended to the context of the \texttt{Optimizer LLM}. By being presented with concrete examples of successful fixes for similar problems, the optimizer is primed to generate a more targeted and higher-quality proposal $\Delta\mathcal{G}_{\text{new}}$. This meta-learning cycle transforms the optimizer from a static component into a dynamic, learning entity, enabling the entire system to learn \emph{how} to improve itself more effectively over time.

\begin{table*}[t!]
\centering
\caption{Performance on the MCP-Universe benchmark, an \textbf{exploratory optimization} scenario. TPGO significantly enhances the success rate of ReAct agents across all four complex task domains by learning from execution feedback.}
\label{tab:main_results}
\begin{tabular}{lccccc}
\toprule
\textbf{Model} & \makecell{\textbf{Repository}\\\textbf{Management}} & \makecell{\textbf{3D}\\\textbf{Designing}} & \makecell{\textbf{Browser}\\\textbf{Automation}} & \makecell{\textbf{Web}\\\textbf{Searching}} & \makecell{\textbf{Overall}\\\textbf{Success Rate}} \\
\midrule
ReAct(GPT-4.1)      & 36.70 & 55.26 & 26.41 & 5.45  & 30.96 \\
+ TPGO(1 iters.) & 40.79 & 52.63 & 27.05 & 8.33 & 32.2\\
+ TPGO(3 iters.) & 41.21 & 60.53 & 32.41 & 12.73 & 36.72\\
+ TPGO(5 iters.) & \textbf{41.21 \color{PosGreen}\scriptsize (+4.51)} & \textbf{65.79 \color{PosGreen}\scriptsize(+10.53)} & \textbf{33.74 \color{PosGreen}\scriptsize(+7.33)} & \textbf{14.55 \color{PosGreen}\scriptsize(+9.10)} & \textbf{38.82 \color{PosGreen}\scriptsize(+7.86)} \\
\midrule
ReAct(DeepSeek-V3.2)      & 42.18 & 60.03 & 20.51 & 16.36  & 34.77 \\
+ TPGO(1 iters.) & 40.85 & 60.52 & 20.38 & 17.14 & 34.72 \\
+ TPGO(3 iters.) & 44.18 & 63.16 & 26.13 & 21.81 & 38.82\\
+ TPGO(5 iters.) & \textbf{45.52 \color{PosGreen}\scriptsize(+3.34)} & \textbf{63.16 \color{PosGreen}\scriptsize(+3.13)} & \textbf{28.03 \color{PosGreen}\scriptsize(+7.52)} & \textbf{30.73 \color{PosGreen}\scriptsize(+14.37)} & \textbf{41.86 \color{PosGreen}\scriptsize(+7.16)} \\
\bottomrule
\end{tabular}
\end{table*}

\begin{table*}[t]
\centering
\caption{Performance on the GAIA validation set, an \textbf{imitative optimization} scenario. TPGO boosts the success rate of the MiroFlow baseline by learning to align with the correct final answer, while also halving the average execution time.}
\label{tab:results}
\begin{tabular}{lccccc}
\toprule
\textbf{Model} & \textbf{Level 1} & \textbf{Level 2} & \textbf{Level 3} & \textbf{Overall} & \textbf{Avg Time (s)} \\
\midrule
MiroFlow (GPT-5)      & 78.8           & 77.5           & 44.4           & 73.8           & 4014 \\
+ TPGO(1 iters.) & 76.9 & 71.2 & 45.5 & 70.59 & 3680\\
+ TPGO(3 iters.) & 82.1 & 73.1 & 54.5 & 75.7 & 2278\\
+ TPGO(5 iters.) & \textbf{90.6} {\color{PosGreen}\scriptsize (+15.0\%)} & \textbf{79.1} {\color{PosGreen}\scriptsize (+2.1\%)} & \textbf{63.6} {\color{PosGreen}\scriptsize (+43.2\%)} & \textbf{81.6} {\color{PosGreen}\scriptsize (+10.6\%)} & \textbf{1765} {\color{blue}\scriptsize (-56.0\%)} \\
\bottomrule
\end{tabular}
\end{table*}

\begin{table}[t]
\centering
\caption{Ablation of key TPGO design choices on the Browser Automation subset of MCP-Universe.}
\label{tab:ablation_components}
\begin{tabular}{lc}
\toprule
\textbf{Method} & \textbf{Success Rate} \\
\midrule
TPGO (full) & \textbf{28.03} \\
w/o Textual Parameter Graph & 25.13 \\
w/o Structural Graph Edits & 24.90 \\
w/o Clustering & 26.52 \\
w/ Random Clustering & 19.97 \\
\bottomrule
\end{tabular}
\end{table}

\section{Experiments}
\label{sec:experiments}

We evaluate TPGO in two representative optimization settings for agent systems: \textbf{exploratory optimization}, where improvement must be driven by execution feedback without a gold trajectory, and \textbf{imitative optimization}, where the system is optimized toward a known correct outcome. Our experiments are designed to answer three questions: (1) whether TPGO consistently improves agent performance in both settings; (2) which components of TPGO are most responsible for these gains; and (3) whether GRAO improves the stability and generalization of iterative optimization. We conduct experiments on MCP-Universe and GAIA, using ReAct and MiroFlow as representative agent frameworks with different backbone models.

\subsection{Experimental Setup}

\paragraph{Benchmarks.}
We evaluate our framework on two challenging benchmarks, each representing one of the core optimization scenarios.
\textbf{MCP-Universe} \cite{luo2025mcp} serves as our testbed for \textit{exploratory optimization}. It is a comprehensive benchmark evaluating LLMs on tasks that require interacting with real-world servers and using large tool spaces. Success is determined by task completion, forcing the agent to rely on self-correction from execution feedback as there is no "golden" path.
\textbf{GAIA} \cite{mialon2023gaia} is chosen for \textit{imitative optimization}. As a widely recognized benchmark for General AI Assistants, it provides tasks with a verifiable "golden" final answer. This allows TPGO to optimize the agent's behavior to align with the desired, correct outcome. We report performance on its validation set across three difficulty levels.

\paragraph{Baseline Systems.}
We apply TPGO to two distinct agent systems to demonstrate its general applicability.
\textbf{ReAct} \cite{DBLP:conf/iclr/YaoZYDSN023} is a foundational agent architecture combining reasoning and acting. We use ReAct agents with both GPT-4.1 and DeepSeek-V3.2 as backbone models on the MCP-Universe benchmark.
\textbf{MiroFlow} \cite{2025mirothinker} is a state-of-the-art open-source framework for building agents capable of complex reasoning. We implement a MiroFlow agent using GPT-5 as the backbone model for the GAIA benchmark.

\paragraph{Implementation Details.}
Unless otherwise specified, the ReAct agents use temperature $=1.0$ and top-$p=1.0$ with GPT-4.1 or DeepSeek-V3.2 as the execution backbone. For MiroFlow on GAIA, the main agent uses GPT-5 with reasoning effort set to \texttt{high}, while sub-agents use GPT-5 with reasoning effort set to \texttt{medium}; both use temperature $=1.0$ and top-$p=1.0$. All TPG parsing, textual-gradient generation, and optimization proposal generation are performed by Gemini-2.5-Pro with temperature $=0.7$ and top-$p=0.95$. For gradient aggregation, we embed negative textual gradients and cluster them using DBSCAN to identify recurring failure patterns without pre-specifying the number of clusters. For each cluster, the optimizer generates a graph-edit proposal, which is then validated on the corresponding task subset before being added to the optimization memory. We run TPGO for up to 5 iterations unless otherwise noted.

\paragraph{Metrics.}
We evaluate performance using two primary metrics.
\textbf{Success Rate (pass@1)} is the percentage of tasks where the system produces the correct final answer in a single attempt, serving as our primary metric for effectiveness.
\textbf{Average Time} is the wall-clock time in seconds from task initiation to completion, measuring system efficiency.

\subsection{Main Results and Analysis}
This section addresses our first research question by comparing the performance of baseline systems with and without the TPGO framework across both optimization scenarios.

\paragraph{Performance in Exploratory Optimization (MCP-Universe)}
As shown in Table \ref{tab:main_results}, applying TPGO in an exploratory setting leads to substantial performance gains. For the GPT-4.1-based ReAct agent, the overall success rate improves from 30.96\% to 38.82\%, a relative increase of 25.4\%. The improvement is most pronounced in the \textit{Web Searching} domain, where the success rate increases from 5.45\% to 14.55\%. This suggests that by learning from execution failures, TPGO effectively identified and rectified systemic flaws in the agent's tool-use logic. Similar gains with the DeepSeek-V3.2 agent, especially in \textit{Web Searching} (+14.37 points), further demonstrate TPGO's ability to optimize agents for diverse and complex workflows based on environmental feedback alone.

\paragraph{Performance in Imitative Optimization (GAIA)}
The results on the GAIA benchmark, presented in Table \ref{tab:results}, demonstrate TPGO's effectiveness in an imitative setting, even when applied to a strong baseline like MiroFlow. By optimizing towards a known correct answer, TPGO improves the overall pass@1 success rate from 73.8\% to 81.6\%, a relative improvement of 10.6\%. Notably, TPGO also significantly enhances system efficiency, reducing the average time per task by 56.0\% (from 4014s to 1765s). This gain stems from the optimization process pruning inefficient reasoning paths and refining agent logic to more directly reach the desired solution. The largest performance gain is on the most difficult (Level 3) tasks, where the success rate increases from 44.4\% to 63.6\%, showing that TPGO improves the system's core capabilities for handling complex problems.

\subsection{Ablation Study}
\label{sec:ablation}

To answer our second research question and identify which design choices are most responsible for TPGO's gains, we conduct a component-level ablation study on the \textbf{Browser Automation} subset of MCP-Universe. Starting from the same ReAct baseline, we selectively disable one design component at a time and report the resulting success rate in Table~\ref{tab:ablation_components}.

The results show that all three components make meaningful contributions. First, removing the \textbf{Textual Parameter Graph} and reverting to monolithic prompt rewriting lowers performance by 2.90 points, confirming that structured modularization is important for effective optimization. Second, disabling \textbf{structural graph edits} causes a 3.13-point drop, indicating that TPGO's gains do not come solely from local text rewriting, but also from the ability to modify agent structure and dependencies. Third, \textbf{clustering} improves the quality of optimization signals: removing clustering reduces performance to 26.52, while replacing semantic clustering with random grouping causes a severe drop to 19.97. This suggests that coherent aggregation of similar failure patterns is crucial, and that noisy grouping can be actively harmful.

Overall, these results indicate that the graph representation, structural edit space, and semantic clustering mechanism are complementary components of TPGO rather than interchangeable implementation details.

\subsection{Optimization Stability Analysis}
\label{sec:stability_analysis}

We further examine whether GRAO improves the \emph{stability} of iterative optimization. On the \textbf{Browser Automation} subset of MCP-Universe, we compare the full \textbf{TPGO (with GRAO)} against \textbf{TPGO (w/o GRAO)}, where the \texttt{Optimizer LLM} updates the policy using only the latest error cluster, without retrieving historical optimization experiences. Both variants start from the same ReAct baseline, and we track success rate over five optimization iterations.

\begin{figure}[t!]
    \setlength{\abovecaptionskip}{5 pt} 
    \setlength{\belowcaptionskip}{-15pt}
    \centering
    \includegraphics[width=1.0\linewidth]{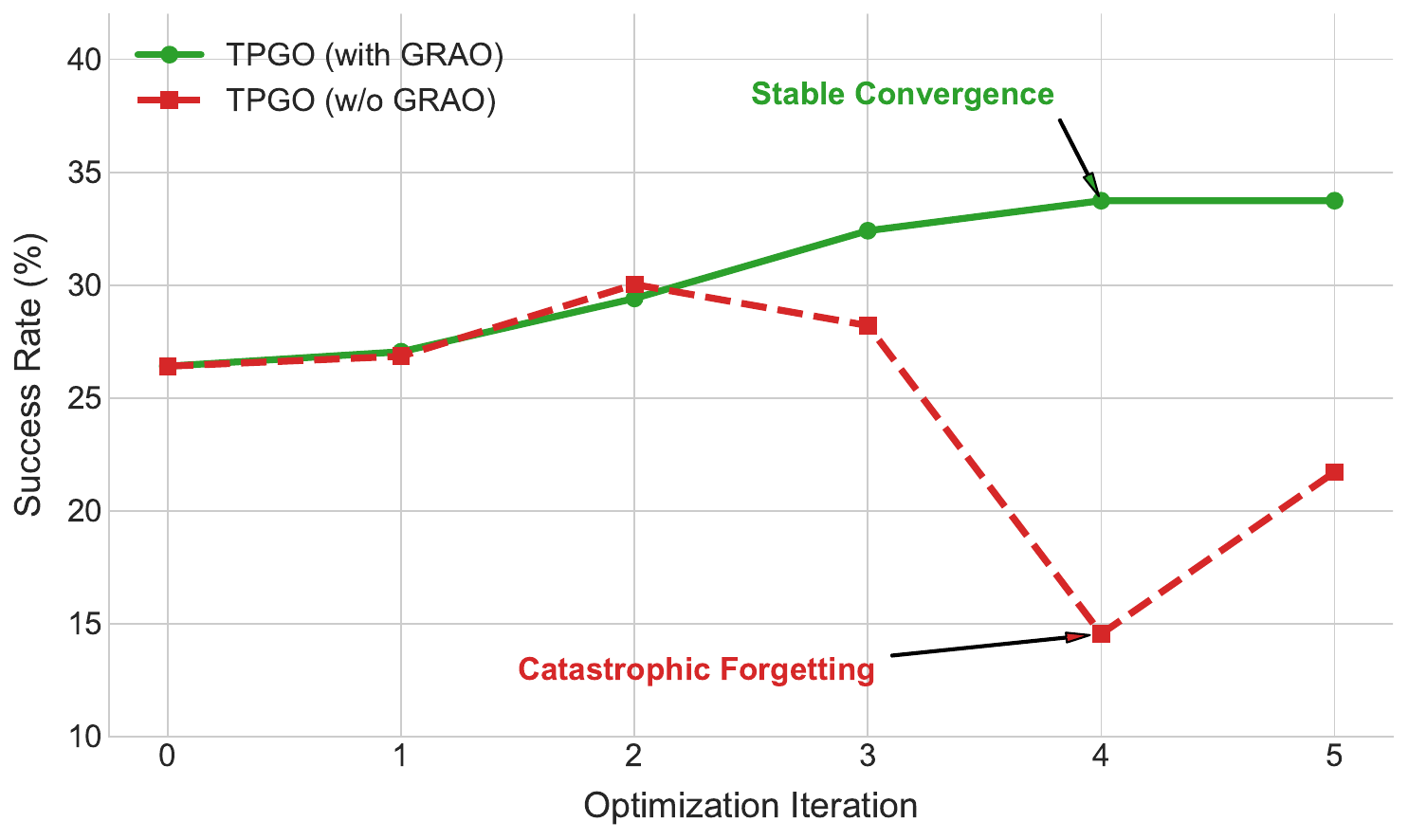}
    \caption{\textbf{Optimization stability with and without GRAO.} Success rate on MCP-Universe Browser Automation over five optimization iterations. TPGO with GRAO improves steadily, while the variant without GRAO suffers from catastrophic forgetting.}
    \label{fig:ablation_grao}
\end{figure}

Figure~\ref{fig:ablation_grao} shows that GRAO is crucial for stable iterative improvement. Both variants start at 26.41\%. With GRAO, TPGO improves steadily to \textbf{33.74\%} by iteration 3 and then remains stable. Without GRAO, performance is much less stable: it peaks at 30.03\% in iteration 2, but then drops sharply to 14.55\% by iteration 4, far below the initial baseline. This suggests that, without historical context, the optimizer overfits to recent failures and introduces updates that hurt previously acquired capabilities. Overall, GRAO improves both final performance and optimization robustness by mitigating catastrophic forgetting.

\subsection{Cross-Domain Generalization}
To examine whether TPGO merely overfits to the task subset used for targeted validation, we conduct a cross-domain generalization study on MCP-Universe. We optimize the DeepSeek-V3.2-based ReAct agent using only the Browser Automation domain and then evaluate the resulting agent on three unseen domains without further optimization. As shown in Table~\ref{tab:cross_domain}, TPGO improves not only the in-domain Browser Automation tasks (+7.52), but also transfers to Web Searching (+1.82), Repository Management (+2.73), and 3D Designing (+3.13). These results suggest that the learned updates capture more general improvements in tool use and agent behavior rather than merely memorizing domain-specific fixes.

\begin{table}[t]
\centering
\small
\caption{Cross-domain generalization on MCP-Universe. TPGO is optimized on Browser Automation only.}
\label{tab:cross_domain}
\begin{tabular}{lcccc}
\toprule
\textbf{Method} & \textbf{Browser} & \textbf{Web} & \textbf{Repo} & \textbf{3D} \\
\midrule
ReAct (DS-V3.2) & 20.51 & 16.36 & 42.18 & 60.03 \\
+ TPGO         & \textbf{28.03} & \textbf{18.18} & \textbf{44.91} & \textbf{63.16} \\
\bottomrule
\end{tabular}
\end{table}

\subsection{Cost-Benefit Analysis}

TPGO introduces additional optimization overhead, so we analyze its cost-benefit trade-off on GAIA. For one full optimization iteration, TPGO consumes 19.9M tokens and 1380 seconds of wall-clock time under 8-way concurrency, corresponding to an amortized cost of 73.7K tokens and 5.6 seconds per trajectory. This one-time offline optimization cost is small relative to the baseline agent's average execution time of 4014 seconds per task. In return, the optimized system improves overall success rate from 73.8\% to 81.6\% and reduces average execution time by 56.0\%. This indicates that TPGO is practical when optimization is amortized over repeated deployment.

\begin{table}[t]
\centering
\caption{Cost-benefit analysis of one TPGO optimization iteration on GAIA.}
\label{tab:cost_benefit}
\begin{tabular}{ll}
\toprule
\textbf{Metric} & \textbf{Value} \\
\midrule
Total token usage & 19.9M \\
Total wall-clock time & 1380s \\
Amortized token usage / trajectory & 73.7K \\
Amortized wall-clock time / trajectory & 5.6s \\
\bottomrule
\end{tabular}
\end{table}

\section{Conclusion}
\label{sec:conclusion}
We introduced Textual Parameter Graph Optimization (TPGO), a framework that improves agent systems through structured graph evolution rather than flat prompt tuning. By decomposing agent configurations into modular textual components, refining them with execution-grounded textual-gradient signals, and leveraging optimization memory through GRAO, TPGO enables more stable and effective agent optimization. Experiments on MCP-Universe and GAIA show consistent gains in both effectiveness and efficiency, suggesting that structured, experience-informed optimization is a promising approach for reducing the manual burden of agent engineering.

\section{Limitations}
TPGO has several limitations. It is an empirical framework without formal optimization guarantees, since ``textual gradients'' are heuristic feedback signals rather than analytical gradients. Its optimization loop can also be costly, as each proposal must be validated on task subsets, which may limit scalability on larger benchmarks or more complex agent systems. In addition, TPGO depends on the capabilities of the underlying LLMs used for parsing, diagnosis, and proposal generation; weaker models may yield noisier feedback or less effective updates. Finally, our current implementation is limited to textual parameter optimization with a predefined set of graph edit operations, and does not yet consider backbone model selection, numerical hyperparameters, or richer structural changes such as node merging and subgraph generation.

\section{Acknowledgments}
This work was supported by Alibaba Group through Alibaba Research Intern Program

\bibliography{custom}

\appendix

\section{LLM usage}
ChatGPT\footnote{https://chat.openai.com/} was used purely with the language of the paper during the writing process, including spell-checking and paraphrasing the authors' original content, without suggesting new content.
Any content generated with the assistant underwent meticulous manual review and subsequently received final approval from the authors.

\section{Ethical Considerations}

No human participants, crowdsourcing, or personally identifiable information (PII) were involved in this research. All experiments were conducted on public benchmarks and standard evaluation splits.

This work studies the automatic optimization of multi-agent systems through textual configuration updates. A potential risk is that improving agent reasoning, tool use, and coordination may also increase the capabilities of such systems in ways that could be misused. To mitigate this concern, we restrict our study to benchmark-based evaluation in controlled settings and do not investigate harmful tasks or safety-sensitive real-world deployment.

In addition, our framework relies on LLM-generated feedback and optimization proposals, which may reflect model errors or biases. Accordingly, the optimized systems should not be interpreted as having formal safety guarantees. We therefore view TPGO as a research framework intended for controlled experimentation, and any practical deployment should include appropriate human oversight and application-specific safeguards.

\section{Implementation Details}

\paragraph{Models and optimization budget.}
We implement TPGO through an OpenAI-compatible API interface. Unless otherwise specified, \texttt{gemini-2.5-pro} is used for all meta-optimization components, including the \textit{Parser LLM}, the \textit{Gradient Generator} (Reflector), and the \textit{Optimizer LLM}. These modules use temperature $0.7$ and top-$p=0.95$. Each API call is retried up to three times with exponential backoff. In all main experiments, we run TPGO for at most five optimization iterations.

\paragraph{Datasets and optimization settings.}
For \textbf{MCP-Universe} \cite{luo2025mcp}, we use the evaluated subset reported in the main results, covering \textbf{146 tasks} from four domains: \textbf{Repository Management}, \textbf{3D Designing}, \textbf{Browser Automation}, and \textbf{Web Searching}. This benchmark is used in an \textit{exploratory optimization} setting, where the Gradient Generator analyzes execution traces without access to gold answers.  
For \textbf{GAIA} \cite{mialon2023gaia}, we start from the public validation split of \textbf{165 questions} and filter it to a \textbf{text-only} subset of \textbf{142 tasks} (Level 1: 49, Level 2: 71, Level 3: 22), matching our focus on textual reasoning and tool use. GAIA is used in an \textit{imitative optimization} setting, where the Gradient Generator additionally receives the ground-truth final answer. To reduce answer leakage, the reflection prompt is instructed to generate generalized behavioral feedback rather than task-specific factual hints.

\paragraph{Graph representation and updates.}
Each agent configuration is represented as a hierarchical JSON-serialized \texttt{PromptNode} structure corresponding to the Textual Parameter Graph (TPG). Nodes are typed as \texttt{role}, \texttt{logic}, or \texttt{tool}; a legacy \texttt{generic} type is retained for compatibility with existing prompts. Graph updates are applied atomically at each iteration by a prompt manager that supports node rewriting, insertion, and deletion.

\paragraph{Textual gradients and clustering.}
For each trajectory $\tau_k$, the Reflector generates negative textual gradients $\delta^-$ that capture failure patterns and positive textual gradients $\delta^+$ that summarize effective behaviors. Negative gradients are embedded and stored in a persistent vector database. We remove near-duplicate entries by cosine-similarity filtering, then cluster the remaining gradients using DBSCAN over \texttt{all-MiniLM-L6-v2} sentence embeddings. In our experiments, DBSCAN uses $\varepsilon = 0.3$ and \texttt{min\_samples} $= 2$. For ablations, we additionally consider variants without clustering and with random clustering.

\paragraph{Optimization memory and rollback.}
GRAO maintains a persistent optimization experience memory. Each entry records the problem context, the proposed graph edit, its rationale, and the observed effectiveness score. When a new error cluster is encountered, candidate experiences are retrieved by semantic similarity and re-ranked by effectiveness; the highest-ranked positive and negative exemplars are then used as in-context demonstrations for the Optimizer LLM. After each iteration, we monitor validation performance and roll back to the previous graph if an update degrades performance on the targeted validation subset. Unsuccessful updates are still retained in memory. For reproducibility, we archive all optimization artifacts, including proposals, graph states, and run metadata, for every iteration.

\input{prompt}

\begin{table*}[htbp]
\centering
\caption{Analysis of a Representative Trajectory Failure}
\label{tab:analysis}
\rowcolors{2}{}{black!5}
\begin{tabularx}{\textwidth}{lX}
\toprule
\rowcolor{tablehead}
\textbf{Component} & \textbf{Description} \\
\midrule
\textbf{Trajectory Flaw} & The agent identified Ollie Watkins but only validated 2 of 4 constraints (current season stats). It neglected to verify historical stats, concluding the task prematurely. \\
\addlinespace
\textbf{Identified Errors} & 
\begin{itemize}[nosep, leftmargin=*]
    \item Concluded the task without systematically verifying all specified constraints.
    \item Focused verification on recent stats while neglecting equally critical historical data.
\end{itemize} \\
\addlinespace
\textbf{Root Cause} & A behavioral tendency to stop investigation once a "good enough" candidate is found, rather than adhering to a strict protocol of validating every single constraint. \\
\bottomrule
\end{tabularx}
\end{table*}

\begin{table*}[htbp]
\centering
\caption{Design of the Corrective Modification}
\label{tab:design}
\rowcolors{2}{}{black!5}
\begin{tabularx}{\textwidth}{lX}
\toprule
\rowcolor{tablehead}
\textbf{Parameter} & \textbf{Rationale \& Implementation} \\
\midrule
\textbf{Action} & A \texttt{new\_node} is created to inject a new, high-priority instruction into the agent's core reasoning process. \\
\addlinespace
\textbf{Content Strategy} & The new instruction establishes a non-skippable "Constraint Validation Protocol." It mandates a clear, multi-step process: (1) List all constraints, (2) Create a checklist, (3) Verify each item with evidence, (4) Proceed only when all items are verified. \\
\addlinespace
\textbf{Rationale} & Creates a mandatory validation checkpoint to prevent premature conclusions and addresses systematic failures in constraint verification by requiring explicit checklist creation. \\
\addlinespace
\textbf{Targeted Errors} & This modification directly resolves the error IDs associated with incomplete constraint validation and premature task conclusion. \\
\bottomrule
\end{tabularx}
\end{table*}

\begin{table*}[htbp]
\centering
\caption{Content of the Resulting Prompt Modification}
\label{tab:implementation}
\rowcolors{2}{}{black!5}
\begin{tabularx}{\textwidth}{lX}
\toprule
\rowcolor{tablehead}
\textbf{Component} & \textbf{Content Detail} \\
\midrule
\textbf{Title} & \texttt{Constraint Validation Protocol} \\
\addlinespace
\textbf{Core Instruction} & A direct command is inserted: "Before providing a final answer, you MUST:"
\begin{enumerate}[nosep, leftmargin=*, label=\arabic*.]
    \item \textbf{List ALL constraints} explicitly.
    \item \textbf{Create a validation checklist}.
    \item \textbf{Verify each constraint} with concrete evidence.
    \item \textbf{Mark each constraint} as \textcolor{darkgreen}{\checkmark VERIFIED} or \textcolor{darkred}{$\times$ UNVERIFIED}.
    \item \textbf{Only proceed} if ALL constraints are verified.
    \item If any are unverified, continue investigation or acknowledge the limitation.
\end{enumerate} \\
\addlinespace
\textbf{Example Format} & A clear example is provided to guide the agent's internal monologue, such as: \texttt{[\checkmark] Constraint 1... [$\times$] Constraint 2...} \\
\addlinespace
\textbf{Rationale} & The node's rationale explicitly states its purpose: "Creates a mandatory validation checkpoint to prevent premature conclusions and addresses systematic failures in constraint verification." \\
\bottomrule
\end{tabularx}
\end{table*}

\section{Case Study}

We present a representative case to illustrate how TPGO diagnoses a trajectory-level failure, maps it to a structured graph update, and produces an interpretable prompt modification. The failure pattern considered here is \textbf{incomplete constraint validation}, a common error mode in which the agent identifies a plausible answer candidate but terminates reasoning before verifying all task requirements.

In the analyzed example, the agent was asked to identify a football player from multiple constraints. Although it correctly proposed Ollie Watkins as the final answer, its reasoning process was incomplete. As shown in Table~\ref{tab:analysis}, the agent verified only a subset of the required conditions, focusing on current-season statistics while omitting historical evidence. The resulting failure is therefore not best characterized as a factual mistake, but as a procedural error: the agent committed to an answer without exhaustively validating all constraints.

TPGO addresses this issue by extracting the underlying behavioral cause from the trajectory rather than only evaluating the correctness of the final output. The analysis in Table~\ref{tab:analysis} attributes the error to a systematic tendency toward premature termination once a sufficiently plausible candidate is found. Such behavior is particularly problematic in agentic settings, where many tasks require explicit satisfaction of multiple constraints and partial verification can lead to confident but unjustified answers.

Based on this diagnosis, TPGO generates a targeted optimization proposal at the graph level. As summarized in Table~\ref{tab:design}, the optimizer introduces a new high-priority node that encodes a \textbf{Constraint Validation Protocol}. Instead of broadly rewriting the full prompt, this update inserts a modular instruction that requires the agent to enumerate all constraints, construct an explicit checklist, verify each item against evidence, and defer answering until all conditions have been checked. This illustrates the central advantage of the Textual Parameter Graph representation: failures can be corrected through localized edits to semantically meaningful components.

The resulting prompt modification is shown in Table~\ref{tab:implementation}. The inserted instruction operationalizes the intended behavioral change by requiring the agent to mark each constraint as verified or unverified before producing a final answer. Notably, this intervention is task-agnostic: it does not inject task-specific facts or encode the correct answer, but instead imposes a reusable reasoning procedure that generalizes to other multi-constraint problems.

Overall, this case exemplifies the full TPGO workflow: trajectory analysis identifies the latent failure pattern, graph-based optimization formulates a structured corrective update, and the final prompt modification instantiates that update as an explicit reasoning protocol. The example also highlights a broader point: in complex agent systems, effective optimization often requires procedural refinement rather than surface-level prompt rewriting. By making these refinements modular and interpretable, TPGO supports both automated optimization and post-hoc inspection.

\end{document}

%% file: prompt.tex
\section{Prompt Templates}
\label{appendix:prompts}

We present the main prompt templates for TPGO’s three LLM modules: Parser, Reflector, and Optimizer. For readability, we summarize the templates, retaining key instructions and output schemas.

\subsection{Graph Construction Prompt (Parser LLM)}
\label{appendix:parser_prompt}

\begin{tcolorbox}[
    title={\textbf{Parser System Prompt}},
    fonttitle=\small\bfseries,
    colback=gray!5, colframe=black!60,
    boxrule=0.5pt, left=3pt, right=3pt, top=3pt, bottom=3pt,
    breakable
]
\footnotesize
\textbf{Role.} Convert an agent prompt into a \textbf{Textual Parameter Graph (TPG)}.

\textbf{Task.} Decompose the input prompt into a hierarchical JSON structure of semantic units for later optimization.

\textbf{Node types.}
\begin{itemize}[leftmargin=1.3em, itemsep=1pt, topsep=2pt]
    \item \textbf{role}: persona, objectives, responsibilities, or high-level directives
    \item \textbf{logic}: reasoning steps, constraints, verification rules, or workflow instructions
    \item \textbf{tool}: tool descriptions, API usage rules, or tool examples
\end{itemize}

\textbf{Key instructions.}
\begin{itemize}[leftmargin=1.3em, itemsep=1pt, topsep=2pt]
    \item Split by semantic function rather than formatting alone.
    \item Parent nodes organize sections; leaf nodes contain actual text.
    \item Do not split atomic instructions, code blocks, JSON objects, or tightly coupled examples.
    \item A node should contain either \texttt{children} or \texttt{content}, but not both.
    \item Preserve the original meaning; do not rewrite or optimize.
\end{itemize}

\textbf{Key output schema.}

{\ttfamily\footnotesize
\{ \\
\hspace*{1em}"title": "<root title>", \\
\hspace*{1em}"type": "generic", \\
\hspace*{1em}"children": [ \\
\hspace*{2em}\{ "title": "<node title>", "type": "<role|logic|tool|generic>", ... \} \\
\hspace*{1em}] \\
\}
}

\textbf{User input.} The model receives the prompt type and the original prompt text, and outputs its TPG decomposition.
\end{tcolorbox}

\subsection{Trajectory Diagnosis Prompt (Reflector LLM)}
\label{appendix:reflection_prompt}

\begin{tcolorbox}[
    title={\textbf{Reflector System Prompt}},
    fonttitle=\small\bfseries,
    colback=gray!5, colframe=black!60,
    boxrule=0.5pt, left=3pt, right=3pt, top=3pt, bottom=3pt,
    breakable
]
\footnotesize
\textbf{Role.} Analyze an execution trajectory and produce \textbf{textual gradients}.

\textbf{Task.}
\begin{itemize}[leftmargin=1.3em, itemsep=1pt, topsep=2pt]
    \item Generate \textbf{negative gradients} ($\delta^-$): abstract failure patterns caused by the agent, such as poor reasoning, missing verification, ineffective coordination, or tool misuse.
    \item Generate \textbf{positive gradients} ($\delta^+$): abstract successful patterns, such as systematic verification, effective tool use, or correct recovery behavior.
\end{itemize}

\textbf{Key instructions.}
\begin{itemize}[leftmargin=1.3em, itemsep=1pt, topsep=2pt]
    \item Describe generalizable behavioral patterns rather than task-specific facts.
    \item Ignore failures caused purely by the environment or external services.
    \item If a reference answer is provided, use it only to diagnose behavior; do not reveal or restate the answer.
\end{itemize}

\textbf{Example abstraction guidance.}\\
Bad: \textit{``The agent ignored that Ollie Watkins had 19 goals.''}\\
Good: \textit{``The agent failed to validate its candidate against all required constraints before producing a final answer.''}

\textbf{Key output schema.}

{\ttfamily\footnotesize
\{ \\
\hspace*{1em}"summary": "<brief behavioral summary>", \\
\hspace*{1em}"error\_list": ["<negative textual gradient>", ...], \\
\hspace*{1em}"experience\_list": ["<positive textual gradient>", ...] \\
\}
}

\textbf{User input.} The model receives the task, the trajectory, and optionally a reference answer.
\end{tcolorbox}

\subsection{Optimization Proposal Prompt (Optimizer LLM)}
\label{appendix:optimizer_prompt}

\begin{tcolorbox}[
    title={\textbf{Optimizer System Prompt}},
    fonttitle=\small\bfseries,
    colback=gray!5, colframe=black!60,
    boxrule=0.5pt, left=3pt, right=3pt, top=3pt, bottom=3pt,
    breakable
]
\footnotesize
\textbf{Role.} Given the current \textbf{Textual Parameter Graph (TPG)} and an error cluster, generate an \textbf{optimization proposal} $\Delta\mathcal{G}$.

\textbf{Inputs.}
\begin{itemize}[leftmargin=1.3em, itemsep=1pt, topsep=2pt]
    \item the current TPG $\mathcal{G}$
    \item an error cluster $C_k$ summarizing a recurring failure pattern
    \item optional historical optimization experiences from \textbf{GRAO}, including similar contexts, graph edits, and effectiveness scores
\end{itemize}

\textbf{Goal.} Propose minimal but effective graph edits that address the root cause while preserving useful behavior.

\textbf{Supported operations.}
\begin{itemize}[leftmargin=1.3em, itemsep=1pt, topsep=2pt]
    \item \texttt{REWRITE\_NODE(v, new\_content)}
    \item \texttt{PRUNE\_EDGE(u, v)}
    \item \texttt{ADD\_NODE(v\_new, content)}
    \item \texttt{ADD\_EDGE(u, v)}
\end{itemize}

\textbf{Key instructions.}
\begin{itemize}[leftmargin=1.3em, itemsep=1pt, topsep=2pt]
    \item Diagnose the root cause before editing.
    \item Prefer reusable fixes over task-specific patches.
    \item If historical experiences are provided, imitate effective strategies and avoid ineffective ones.
    \item Keep the proposal internally consistent with the current graph.
\end{itemize}

\textbf{Key output schema.}

{\ttfamily\footnotesize
\{ \\
\hspace*{1em}"problem\_context": "<abstract failure pattern>", \\
\hspace*{1em}"modifications": [ \\
\hspace*{2em}\{ \\
\hspace*{3em}"operation": "REWRITE\_NODE | PRUNE\_EDGE | ADD\_NODE | ADD\_EDGE", \\
\hspace*{3em}"target": \{ ... \}, \\
\hspace*{3em}"new\_node": \{ ... \}, \\
\hspace*{3em}"new\_content": "<replacement content if applicable>", \\
\hspace*{3em}"addresses\_errors": [<error indices>], \\
\hspace*{3em}"rationale": "<why this edit helps>" \\
\hspace*{2em}\} \\
\hspace*{1em}] \\
\}
}

\textbf{User input.} The model receives the current TPG, the error cluster, and optional historical optimization experiences.
\end{tcolorbox}

%% file: custom.bib
@article{wang2023promptagent,
  title={Promptagent: Strategic planning with language models enables expert-level prompt optimization},
  author={Wang, Xinyuan and Li, Chenxi and Wang, Zhen and Bai, Fan and Luo, Haotian and Zhang, Jiayou and Jojic, Nebojsa and Xing, Eric P and Hu, Zhiting},
  journal={arXiv preprint arXiv:2310.16427},
  year={2023}
}

@inproceedings{tong2025evoprompt,
  title={Evoprompt: Evolving prompts for enhanced zero-shot named entity recognition with large language models},
  author={Tong, Zeliang and Ding, Zhuojun and Wei, Wei},
  booktitle={Proceedings of the 31st International Conference on Computational Linguistics},
  pages={5136--5153},
  year={2025}
}

@article{yuksekgonul2024textgrad,
  title={Textgrad: Automatic" differentiation" via text},
  author={Yuksekgonul, Mert and Bianchi, Federico and Boen, Joseph and Liu, Sheng and Huang, Zhi and Guestrin, Carlos and Zou, James},
  journal={arXiv preprint arXiv:2406.07496},
  year={2024}
}

@inproceedings{DBLP:conf/iclr/YaoZYDSN023,
  author       = {Shunyu Yao and
                  Jeffrey Zhao and
                  Dian Yu and
                  Nan Du and
                  Izhak Shafran and
                  Karthik R. Narasimhan and
                  Yuan Cao},
  title        = {ReAct: Synergizing Reasoning and Acting in Language Models},
  booktitle    = {The Eleventh International Conference on Learning Representations,
                  {ICLR} 2023, Kigali, Rwanda, May 1-5, 2023},
  publisher    = {OpenReview.net},
  year         = {2023},
  url          = {https://openreview.net/forum?id=WE\_vluYUL-X},
  bibsource    = {dblp computer science bibliography, https://dblp.org}
}

@inproceedings{DBLP:conf/ijcai/GuoCWCPCW024,
  author       = {Taicheng Guo and
                  Xiuying Chen and
                  Yaqi Wang and
                  Ruidi Chang and
                  Shichao Pei and
                  Nitesh V. Chawla and
                  Olaf Wiest and
                  Xiangliang Zhang},
  title        = {Large Language Model Based Multi-agents: {A} Survey of Progress and
                  Challenges},
  booktitle    = {Proceedings of the Thirty-Third International Joint Conference on
                  Artificial Intelligence, {IJCAI} 2024, Jeju, South Korea, August 3-9,
                  2024},
  pages        = {8048--8057},
  publisher    = {ijcai.org},
  year         = {2024},
  url          = {https://www.ijcai.org/proceedings/2024/890},
  bibsource    = {dblp computer science bibliography, https://dblp.org}
}

@inproceedings{DBLP:conf/uist/ParkOCMLB23,
  author       = {Joon Sung Park and
                  Joseph C. O'Brien and
                  Carrie Jun Cai and
                  Meredith Ringel Morris and
                  Percy Liang and
                  Michael S. Bernstein},
  editor       = {Sean Follmer and
                  Jeff Han and
                  J{\"{u}}rgen Steimle and
                  Nathalie Henry Riche},
  title        = {Generative Agents: Interactive Simulacra of Human Behavior},
  booktitle    = {Proceedings of the 36th Annual {ACM} Symposium on User Interface Software
                  and Technology, {UIST} 2023, San Francisco, CA, USA, 29 October 2023-
                  1 November 2023},
  pages        = {2:1--2:22},
  publisher    = {{ACM}},
  year         = {2023},
  url          = {https://doi.org/10.1145/3586183.3606763},
  doi          = {10.1145/3586183.3606763},
  bibsource    = {dblp computer science bibliography, https://dblp.org}
}

@article{zhao2024mactg,
  title={MaCTG: Multi-Agent Collaborative Thought Graph for Automatic Programming},
  author={Zhao, Zixiao and Sun, Jing and Hou, Zhe and Wei, Zhiyuan and Cai, Cheng-Hao and Qiao, Miao and Dong, Jin Song},
  journal={arXiv preprint arXiv:2410.19245},
  year={2024}
}

@article{lin2025creativity,
  title={Creativity in LLM-based Multi-Agent Systems: A Survey},
  author={Lin, Yi-Cheng and Chen, Kang-Chieh and Li, Zhe-Yan and Wu, Tzu-Heng and Wu, Tzu-Hsuan and Chen, Kuan-Yu and Lee, Hung-yi and Chen, Yun-Nung},
  journal={arXiv preprint arXiv:2505.21116},
  year={2025}
}

@inproceedings{DBLP:conf/acl/ZhangX0LHD24,
  author       = {Jintian Zhang and
                  Xin Xu and
                  Ningyu Zhang and
                  Ruibo Liu and
                  Bryan Hooi and
                  Shumin Deng},
  editor       = {Lun{-}Wei Ku and
                  Andre Martins and
                  Vivek Srikumar},
  title        = {Exploring Collaboration Mechanisms for {LLM} Agents: {A} Social Psychology
                  View},
  booktitle    = {Proceedings of the 62nd Annual Meeting of the Association for Computational
                  Linguistics (Volume 1: Long Papers), {ACL} 2024, Bangkok, Thailand,
                  August 11-16, 2024},
  pages        = {14544--14607},
  publisher    = {Association for Computational Linguistics},
  year         = {2024},
  url          = {https://doi.org/10.18653/v1/2024.acl-long.782},
  doi          = {10.18653/V1/2024.ACL-LONG.782},
  bibsource    = {dblp computer science bibliography, https://dblp.org}
}

@inproceedings{mialon2023gaia,
  title={Gaia: a benchmark for general ai assistants},
  author={Mialon, Gr{\'e}goire and Fourrier, Cl{\'e}mentine and Wolf, Thomas and LeCun, Yann and Scialom, Thomas},
  booktitle={The Twelfth International Conference on Learning Representations},
  year={2023}
}

@article{luo2025mcp,
  title={Mcp-universe: Benchmarking large language models with real-world model context protocol servers},
  author={Luo, Ziyang and Shen, Zhiqi and Yang, Wenzhuo and Zhao, Zirui and Jwalapuram, Prathyusha and Saha, Amrita and Sahoo, Doyen and Savarese, Silvio and Xiong, Caiming and Li, Junnan},
  journal={arXiv preprint arXiv:2508.14704},
  year={2025}
}

@misc{2025mirothinker,
    title={MiroFlow: A High-Performance Open-Source Research Agent Framework},
    author={MiroMind AI Team},
    howpublished={\url{https://github.com/MiroMindAI/MiroFlow}},
    year={2025}
}

@inproceedings{hong2023metagpt,
  title={MetaGPT: Meta programming for a multi-agent collaborative framework},
  author={Hong, Sirui and Zhuge, Mingchen and Chen, Jonathan and Zheng, Xiawu and Cheng, Yuheng and Wang, Jinlin and Zhang, Ceyao and Wang, Zili and Yau, Steven Ka Shing and Lin, Zijuan and others},
  booktitle={The Twelfth International Conference on Learning Representations},
  year={2023}
}

@article{talebirad2023multi,
  title={Multi-agent collaboration: Harnessing the power of intelligent llm agents},
  author={Talebirad, Yashar and Nadiri, Amirhossein},
  journal={arXiv preprint arXiv:2306.03314},
  year={2023}
}

@article{luo2025large,
  title={Large language model agent: A survey on methodology, applications and challenges},
  author={Luo, Junyu and Zhang, Weizhi and Yuan, Ye and Zhao, Yusheng and Yang, Junwei and Gu, Yiyang and Wu, Bohan and Chen, Binqi and Qiao, Ziyue and Long, Qingqing and others},
  journal={arXiv preprint arXiv:2503.21460},
  year={2025}
}

@article{tran2025multi,
  title={Multi-agent collaboration mechanisms: A survey of llms},
  author={Tran, Khanh-Tung and Dao, Dung and Nguyen, Minh-Duong and Pham, Quoc-Viet and O'Sullivan, Barry and Nguyen, Hoang D},
  journal={arXiv preprint arXiv:2501.06322},
  year={2025}
}

@article{fang2025comprehensive,
  title={A comprehensive survey of self-evolving ai agents: A new paradigm bridging foundation models and lifelong agentic systems},
  author={Fang, Jinyuan and Peng, Yanwen and Zhang, Xi and Wang, Yingxu and Yi, Xinhao and Zhang, Guibin and Xu, Yi and Wu, Bin and Liu, Siwei and Li, Zihao and others},
  journal={arXiv preprint arXiv:2508.07407},
  year={2025}
}
